\documentclass{article}

\usepackage[preprint]{neurips_2026}
\usepackage[utf8]{inputenc} % allow utf-8 input
\usepackage[T1]{fontenc}    % use 8-bit T1 fonts
\usepackage{hyperref}       % hyperlinks
\usepackage{url}            % simple URL typesetting
\usepackage{booktabs}       % professional-quality tables
\usepackage{amsfonts}       % blackboard math symbols
\usepackage{nicefrac}       % compact symbols for 1/2, etc.
\usepackage{microtype}      % microtypography
\usepackage{xcolor}         % colors
\usepackage{graphicx}
\usepackage{svg}
\usepackage{amsmath}
\title{GSToken: Geometry-Structured Gaussian Tokens for Compact 3D Medical Image Representation}

\author{
  Xiaoduo Li \\
  Taiyuan University of Technology \\
  \texttt{2024001756@link.tyut.edu.cn} \\
  \And
  Quan Gu \\
  independent researcher \\
  \texttt{guquan1839@163.com}
}

\begin{document}

\maketitle

\begin{abstract}
Effective segmentation of multi-modal MRI is central to improving neural network accuracy in brain tumor recognition. Existing methods typically compress 3D volumes into token sequences via fixed patch encoding or learned attention pooling (e.g., TokenLearner). However, these compression schemes discard explicit spatial shape information—the resulting tokens convey no notion of lesion morphology or spatial extent. Meanwhile, end-to-end evaluation entangles a tokenizer's information retention with the reconstruction capacity of the downstream decoder, and the lack of a unified capacity contract across methods makes performance differences difficult to attribute. In this paper, we introduce Gaussian tokens to multi-modal brain tumor segmentation for the first time: each token carries not only a semantic feature but also a learned 3D center, anisotropic scale, and orientation, endowing the representation with explicit geometric support at negligible parameter cost. We further propose a frozen-token utility evaluation protocol—the trained tokenizer is frozen, its output is cast into a fixed-capacity serialized contract, and a shared lightweight Transformer probe independently measures each tokenizer's retained information under strictly matched conditions. Multi-seed paired statistical testing shows that GSToken consistently and substantially outperforms capacity-matched adaptive baselines under frozen probing, with uniform advantages across all tumor sub-regions, surface, and distance metrics. These results demonstrate that explicitly encoding spatial geometry within tokens significantly improves the information density of volumetric representations, offering a new design principle for compact 3D medical image representation and downstream reading.
\end{abstract}

\section{Introduction}

Automatic segmentation of three-dimensional magnetic resonance imaging (MRI) is essential for visualizing lesions, planning treatment, and monitoring therapeutic response in neuro-oncology~\cite{baid2021rsnaasnrmiccaibrats2021benchmark}. However, 3D MRI volumes are dense, high-resolution grids, and Transformer-based segmentation models built on the U-Net family~\cite{10.1007/978-3-319-24574-4_28,7785132,10.1007/978-3-319-46723-8_49,oktay2018attentionunetlearninglook,10.1007/978-3-030-00889-5_1,alom2018recurrentresidualconvolutionalneural,isensee2018nnunetselfadaptingframeworkunetbased} require compressing them into token sequences; direct voxel tokenization produces sequences too long for practical computation, while the tokenization itself determines how much task-relevant information survives downstream~\cite{dosovitskiy2021an,chen2021transunettransformersmakestrong,Hatamizadeh_2022_WACV,ryoo2022tokenlearner8learnedtokens}. The core question is therefore not \emph{whether} volume data can be compressed, but \emph{how much} segmentation-relevant information a fixed token budget can preserve.

Existing tokenization schemes fall into two families. \textbf{Fixed-grid patch tokens} divide the volume into regular non-overlapping patches (ViT~\cite{dosovitskiy2021an}, TransUNet~\cite{chen2021transunettransformersmakestrong}, UNETR~\cite{Hatamizadeh_2022_WACV}); they are simple and uniform but their geometry is decoupled from lesion morphology, and small or anisotropic structures such as enhancing tumor boundaries are easily diluted across patches. \textbf{Learned adaptive tokens} replace the grid with data-dependent aggregation (TokenLearner~\cite{ryoo2022tokenlearner8learnedtokens}, Perceiver~\cite{pmlr-v139-jaegle21a}), improving flexibility, yet the resulting tokens carry no explicit 3D support --- no position, extent, or orientation that encodes where and how a structure extends in space. In parallel, Gaussian primitives have proven a powerful explicit representation in 3D scene modeling: 3D Gaussian Splatting (3DGS)~\cite{10.1145/3592433} achieves high-quality rendering with anisotropic Gaussians, and recent works extend Gaussian primitives to medical reconstruction, interactive segmentation, and volume visualization~\cite{liang2026innergsinternalscenesreconstruction,Kumar_2025,jeon2026supergaussianinteractivesceneediting}. Despite their expressive geometry, these Gaussian representations have not been exploited as general-purpose tokens for segmentation-oriented representation learning, and to our knowledge, no study isolates whether Gaussian-structured tokens preserve more segmentation information than grid or learned-pooling tokens under strictly matched capacity.

We propose \textbf{GSToken}, which tokenizes 3D multi-modal MRI into adaptive Gaussian tokens --- each token jointly encodes a content feature with an explicit anisotropic support described by center, scale, and rotation. A 3D convolutional encoder produces dense features; a Gaussian proposal module predicts token parameters; and a lesion-aware allocation scheme re-distributes the fixed token budget toward complex tumor regions. Serialized Gaussian tokens are then read by a unified Transformer reader that reconstructs whole-tumor (WT), tumor-core (TC), and enhancing-tumor (ET) segmentation. To isolate the value of the representation itself, we adopt a strict capacity-matching protocol --- identical token count, feature width, serialized payload, reader architecture, and training budget --- comparing GSToken against Patch and TokenLearner baselines on BraTS 2021~\cite{baid2021rsnaasnrmiccaibrats2021benchmark}.

\begin{figure}[htbp]
    \centering
    \includegraphics[width=\linewidth]{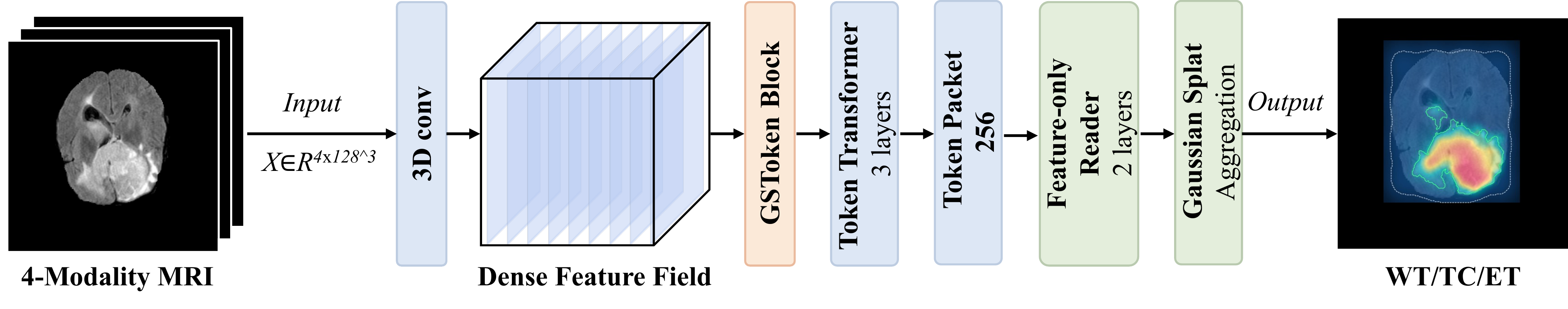}
    \caption{Overall framework of the proposed GSToken method.}
    \label{fig:framework1}
\end{figure}

\begin{figure}[htbp]
    \centering
    \includegraphics[width=0.8\linewidth]{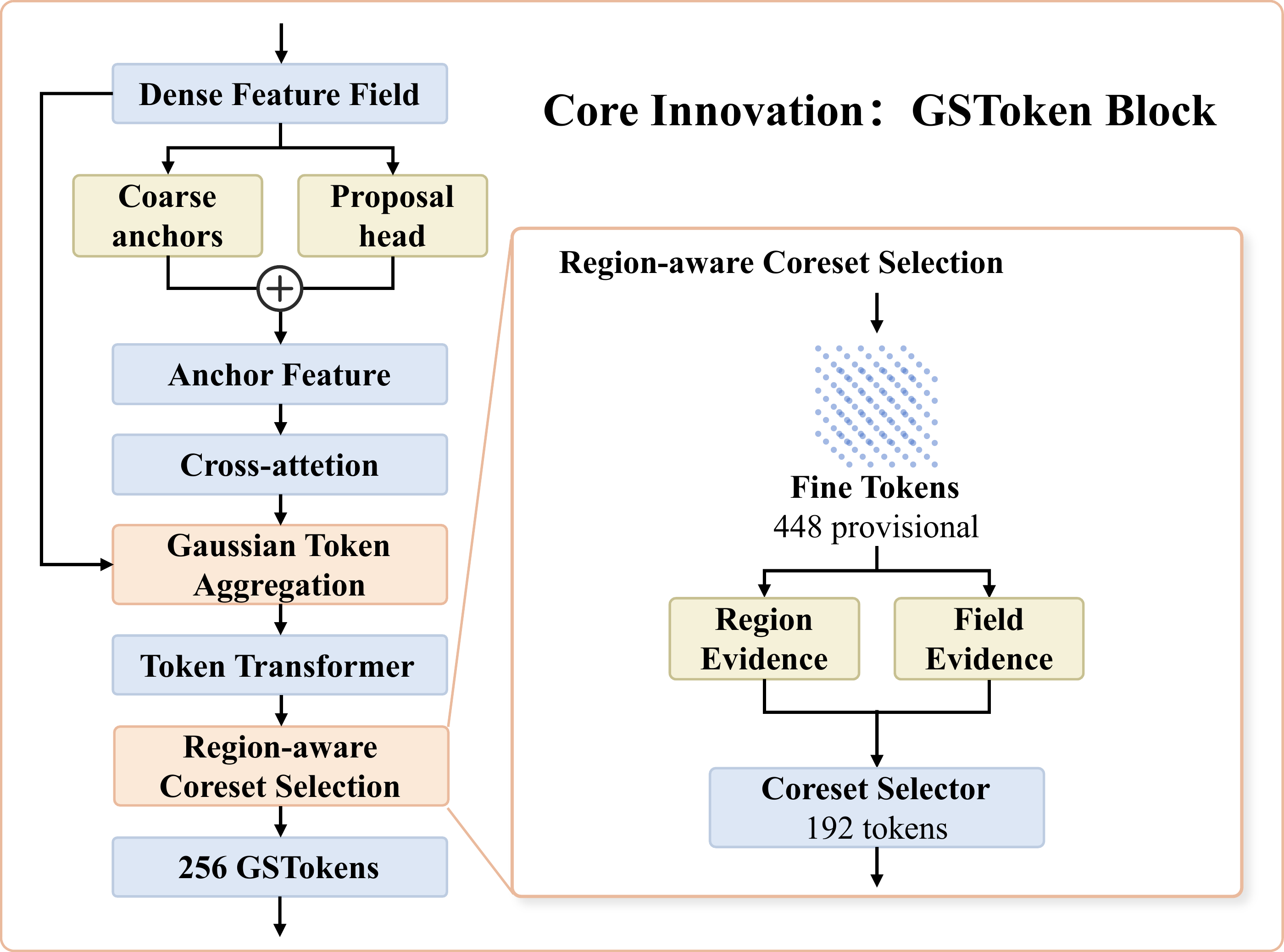}
    \caption{Framework of the GSToken block.}
    \label{fig:framework2}
\end{figure}

Our contributions are: (i) we introduce geometry-structured adaptive Gaussian tokens for 3D MRI, representing lesion content and spatial support jointly under a fixed token budget; (ii) we establish a frozen-token evaluation protocol that measures tokenizer-readability in isolation from downstream decoder capacity; and (iii) we show that GSToken consistently outperforms capacity-matched Patch and TokenLearner baselines across three seeds (Macro Dice +0.221 $\pm$ 0.035, 3/3 seeds favorable on WT/TC/ET Dice, Surface Dice, and HD95), demonstrating that explicit Gaussian geometry improves information preservation in compact medical representations.

\section{Related Work}
\paragraph{3D Medical Image Segmentation}
To capture long-range dependencies in volumetric data, Transformers have been progressively introduced into 3D medical image segmentation. TransBTS integrates a Transformer into a 3D CNN framework~\cite{10.1007/978-3-030-87193-2_11}. Its 3D CNN encoder first extracts local voxel-level features and reshapes the resulting feature maps into a sequence of tokens. A Transformer module then captures long-range dependencies, after which a decoder progressively upsamples the features to recover high-resolution segmentation maps. TransBTS achieves performance comparable to or better than previous state-of-the-art methods on the BraTS 2019 and 2020 datasets. In the same year, DiNTS introduced differentiable neural architecture search into 3D medical image segmentation~\cite{He_2021_CVPR}. It jointly searches for network topologies and operations within a flexible multi-path search space, employs a topology loss to bridge the gap between continuous optimization and the resulting discrete architecture, and constrains the search according to a predefined GPU memory budget. DiNTS achieves state-of-the-art performance across all ten tasks of the Medical Segmentation Decathlon.

Given the high computational cost of Transformers, alternative architectures have also emerged. For example, 3D UX-Net replaces Swin-style self-attention with large-kernel depthwise convolutions and uses pointwise convolutions for channel mixing~\cite{lee2023d}. This design introduces the inductive bias and parameter efficiency of convolutional networks while maintaining a large receptive field. It consistently outperforms the evaluated Transformer-based methods on the FLARE2021~\cite{MA2022102616}, FeTA2021~\cite{Payette_2023}, and AMOS2022~\cite{ji2022amoslargescaleabdominalmultiorgan} datasets. Taken together, these studies demonstrate that balancing global context modeling with computational efficiency has become a central direction in 3D medical image segmentation.

\paragraph{Visual Tokenization and Token Reduction}
Discretizing continuous images into tokens provides a common foundation for generative and self-supervised vision models. VQ-VAE maps continuous encoder features to discrete latent codes through a vector-quantized codebook and enables end-to-end training using a straight-through estimator~\cite{3295222.3295378}. This approach mitigates the posterior-collapse problem commonly observed in continuous latent-variable models and allows autoregressive models such as PixelCNN to model the learned discrete representations, thereby laying the foundation for visual tokenization. Building on this framework, VQGAN incorporates adversarial training and perceptual losses into vector-quantized representation learning to substantially improve reconstruction quality~\cite{Esser2020TamingTF}. The resulting discrete tokens can then be modeled by an autoregressive Transformer to generate high-quality images. Discrete visual tokens also serve as prediction targets for masked image pretraining. BEiT adapts BERT-style masked language modeling to vision by randomly masking image patches and predicting discrete visual tokens generated by a pretrained dVAE tokenizer~\cite{bao2022beit}. In contrast, MAE masks 75\% of image patches and reconstructs their raw pixels, encouraging the model to infer global semantic structure from the visible patches~\cite{He2021MaskedAA}. This simple strategy has become a dominant paradigm in self-supervised visual pretraining.

For efficient inference, the redundancy among visual tokens has motivated extensive research on token reduction. One line of work directly removes unimportant tokens. DynamicViT uses lightweight prediction modules to estimate token importance and progressively discards less informative tokens during inference~\cite{3540261.3541329}. It provides early systematic evidence that token sparsification can substantially reduce the computational cost of Vision Transformers while largely preserving accuracy. EViT identifies informative image tokens using class-token attention, preserves the most relevant tokens, and fuses the remaining tokens into a single representative token~\cite{Liang2022NotAP}. Its parameter-free token reorganization mechanism is jointly optimized with the backbone and accelerates inference with only a minor loss in accuracy. Another line of work reduces redundancy by merging similar tokens rather than discarding them. ToMe performs token merging through bipartite soft matching and can substantially accelerate Vision Transformers without additional training, typically with only a negligible reduction in accuracy~\cite{bolya2023token}. It has subsequently become a widely adopted baseline for token-merging research. Together, these studies motivate compact token representations that preserve spatial structure while reducing sequence length.

\section{Gaussian Token (GSToken)}
\label{sec:gstoken}

This section introduces the representation of the proposed Gaussian token and
describes how it is integrated into the overall network. We further present the
adaptive coarse-to-fine allocation, region-aware token selection, and optional
prediction refinement used to improve the representation of locally salient
structures.

\subsection{Gaussian Token Representation}
\label{sec:gs_token_representation}

We initialize $K_c$ coarse tokens on a fixed regular grid to maintain global
spatial coverage and generate $K_p$ provisional fine tokens adaptively from
each input volume to represent locally salient structures. The region-aware
selector later retains $K_f$ fine tokens, giving a final token budget of
$K=K_c+K_f$. When token selection is disabled, $K_f=K_p$.

Each Gaussian token is represented by the five-tuple
\begin{equation}
g_i =
\left(
\boldsymbol{\mu}_i,
\boldsymbol{s}_i,
\mathbf{R}_i,
\alpha_i,
\mathbf{t}_i
\right),
\label{eq:gstoken_definition}
\end{equation}
where $\boldsymbol{\mu}_i\in[-1,1]^3$ is the Gaussian center,
$\boldsymbol{s}_i\in\mathbb{R}_{+}^{3}$ denotes its anisotropic scale,
$\mathbf{R}_i\in SO(3)$ is its rotation matrix, $\alpha_i\in(0,1]$ denotes
its importance, and $\mathbf{t}_i\in\mathbb{R}^{C}$ is its content
feature.

The center $\boldsymbol{\mu}_i$ is obtained by adding a bounded learnable
offset to the corresponding anchor $\mathbf{a}_i\in[-1,1]^3$. Coarse anchors
are defined on a regular grid, while fine anchors are generated by the adaptive
proposal mechanism described below. The scale $\boldsymbol{s}_i$ is
parameterized relative to a resolution-dependent reference scale: the coarse
grid resolution is used for coarse tokens, whereas the encoded feature-field
resolution is used for fine tokens. This reference scale is modulated by a
learned logarithmic correction. The geometry head additionally predicts a quaternion
$\boldsymbol{\rho}_i\in\mathbb{R}^{4}$, which is normalized and converted into
$\mathbf{R}_i$ to determine the orientation of the Gaussian ellipsoid. The
importance $\alpha_i$ modulates the contribution of the token during Gaussian
splatting. The quaternion $\boldsymbol{\rho}_i$ is used only to parameterize
$\mathbf{R}_i$ and is therefore not stored as an additional token attribute.

Let $\mathbf{F}\in\mathbb{R}^{C\times H'\times W'\times D'}$ be the dense
feature field produced by the encoder, and let $V=H'W'D'$. We denote its
feature and normalized coordinate at location $v$ by
$\mathbf{F}_v\in\mathbb{R}^{C}$ and $\mathbf{x}_v\in[-1,1]^3$, respectively.
The spatial support of token $i$ at location $v$ is defined by the anisotropic
Gaussian log-weight
\begin{equation}
\ell_{iv}
=
-\frac{1}{2}
\left\|
\mathbf{R}_i^{\top}
\left(\mathbf{x}_v-\boldsymbol{\mu}_i\right)
\oslash
\boldsymbol{s}_i
\right\|_2^2,
\label{eq:gaussian_log_weight}
\end{equation}
where $\oslash$ denotes element-wise division. Equivalently, defining
$\boldsymbol{\Sigma}_i=\mathbf{R}_i
\operatorname{diag}(\boldsymbol{s}_i^2)\mathbf{R}_i^{\top}$, the same score
can be expressed as
\begin{equation}
\ell_{iv}
=
-\frac{1}{2}
\left(\mathbf{x}_v-\boldsymbol{\mu}_i\right)^{\top}
\boldsymbol{\Sigma}_i^{-1}
\left(\mathbf{x}_v-\boldsymbol{\mu}_i\right).
\label{eq:gaussian_mahalanobis}
\end{equation}
Here, $\ell_{iv}$ measures the spatial compatibility between token $i$ and
location $v$: locations close to the Gaussian center and aligned with its
principal axes receive larger values, while distant locations receive smaller
values.

To gather a feature for token $i$, the log-weights are normalized over the $V$
locations of the dense feature field:
\begin{equation}
A_{iv}^{\mathrm{gather}}
=
\frac{\exp(\ell_{iv})}
{\sum_{u=1}^{V}\exp(\ell_{iu})},
\qquad
\mathbf{z}_i
=
\sum_{v=1}^{V}
A_{iv}^{\mathrm{gather}}\mathbf{F}_v,
\label{eq:gaussian_gathering}
\end{equation}
where $\mathbf{z}_i\in\mathbb{R}^{C}$ denotes the gathered feature before token
interaction. Thus, each token aggregates information from a continuous,
anisotropic, and orientation-aware support region rather than from a fixed point
or cubic patch.

The gathered feature $\mathbf{z}_i$ is fused with the cross-attention context
$\mathbf{h}_i$ and processed by the token interaction module, producing the
token content $\mathbf{t}_i$ used in Eq.~\eqref{eq:gstoken_definition}. The same
Gaussian geometry is then used in the reverse direction to splat these token
features back to the dense field:
\begin{equation}
A_{iv}^{\mathrm{splat}}
=
\frac{\exp\!\left(\ell_{iv}+\log(\alpha_i+\epsilon)\right)}
{\sum_{j=1}^{K}\exp\!\left(\ell_{jv}+\log(\alpha_j+\epsilon)\right)},
\qquad
\widehat{\mathbf{F}}_v
=
\sum_{i=1}^{K}
A_{iv}^{\mathrm{splat}}\mathbf{t}_i.
\label{eq:gaussian_splatting}
\end{equation}
Unlike gathering, which normalizes over spatial locations for each token,
splatting normalizes over tokens for each location. Consequently, the learned
center, scale, and rotation directly control both token formation and dense
feature reconstruction. Here, $\epsilon>0$ is a small constant used for
numerical stability, and $A_{iv}^{\mathrm{gather}}$ and
$A_{iv}^{\mathrm{splat}}$ denote the normalized gathering and splatting
weights, respectively.

\subsection{Overall Network Architecture}
\label{sec:overall_architecture}

As illustrated in Fig.~\ref{fig:framework1}, the 3D convolutional encoder
$E_{\theta}$ first maps the input $\mathbf{X}$ to the dense feature field
$\mathbf{F}$. The proposal head
and the regular coarse grid jointly generate $K_p$ provisional fine anchors and
$K_c$ coarse anchors, respectively. For each anchor, a local feature is sampled
from $\mathbf{F}$ and combined with a learnable query
$\mathbf{q}_i\in\mathbb{R}^{C}$. Cross-attention with the flattened feature
field produces a contextual feature $\mathbf{h}_i\in\mathbb{R}^{C}$ for
predicting the Gaussian center, scale, and rotation.

The predicted geometry defines the Gaussian gathering weights in
Eq.~\eqref{eq:gaussian_gathering}, producing a provisional set
$\mathcal{G}_{\mathrm{prov}}$ of $K_c+K_p$ spatially grounded tokens. Their
gathered features $\mathbf{z}_i$ are fused with the cross-attention context
$\mathbf{h}_i$ and processed by an $L$-layer token Transformer $\Phi_L$. Its
outputs are the token content features $\mathbf{t}_i$. The Transformer models
interactions across spatial locations and between coarse and fine tokens. A
region-aware coreset selector then
retains all coarse tokens and selects $K_f$ complementary fine tokens, yielding
the final set $\mathcal{G}=\{g_i\}_{i=1}^{K}$ with $K=K_c+K_f$.

In the default dense-prediction pathway, the transformed token features are
splat back to a dense feature field using Eq.~\eqref{eq:gaussian_splatting},
and a 3D decoder $D_{\psi}$ produces the final prediction
$\widehat{\mathbf{Y}}$. Here, $\mathcal{S}_G$ denotes the Gaussian splatting
operator defined in Eq.~\eqref{eq:gaussian_splatting},
$\widehat{\mathbf{F}}$ is the reconstructed dense feature field, and
$D_{\psi}$ is the dense prediction decoder. The complete pathway is summarized
as
\begin{equation}
\mathbf{X}
\xrightarrow{E_{\theta}}
\mathbf{F}
\xrightarrow{\text{allocation and gathering}}
\mathcal{G}_{\mathrm{prov}}
\xrightarrow{\Phi_L}
\widetilde{\mathcal{G}}_{\mathrm{prov}}
\xrightarrow{\text{coreset selection}}
\mathcal{G}
\xrightarrow{\mathcal{S}_G}
\widehat{\mathbf{F}}
\xrightarrow{D_{\psi}}
\widehat{\mathbf{Y}}.
\label{eq:overall_pipeline}
\end{equation}
In Eq.~\eqref{eq:overall_pipeline}, $\mathcal{G}_{\mathrm{prov}}$ denotes the
provisional set of coarse and fine tokens,
$\widetilde{\mathcal{G}}_{\mathrm{prov}}$ denotes the same set after token
interaction, and $\mathcal{G}$ denotes the final fixed-budget subset after
coreset selection.

To evaluate whether the compact representation itself preserves sufficient
information, we additionally use an optional feature-only reader. As shown by
the ``Token Packet--Feature-only Reader--Gaussian Splat Aggregation'' branch in
Fig.~\ref{fig:framework1}, this reader receives only the final token packet and
does not access the input image or dense encoder features. It is used solely as
an evaluation protocol and is not required by the default prediction network.

\subsection{Adaptive GSToken Construction and Optimization}
\label{sec:adaptive_token_construction}

\subsubsection{Coarse-to-Fine Candidate Allocation}
\label{sec:coarse_to_fine}

A regular token grid assigns equal representation capacity to every spatial
region, although informative structures may occupy only a small portion of a
volume. We therefore combine two complementary anchor sets. The $K_c$ coarse
anchors are placed on a fixed regular grid and preserve global coverage. A
lightweight proposal head predicts a spatial importance map from the dense
feature field and uses it to generate $K_p$ image-adaptive fine anchors. In the
dynamic variant, proposal confidence can be combined with coarse-reconstruction
residuals and region-related evidence to increase candidate recall in
underrepresented areas.

For each coarse or fine anchor, the network samples a local anchor feature and
adds it to a learnable query. The query attends to the flattened dense feature
field through cross-attention, and the resulting context is used to predict the
Gaussian geometry. The learned geometry then gathers the token content from
the dense feature field. Finally, the gathered content is fused with the
attention context and passed through the token Transformer. This procedure
produces the provisional set of $K_c+K_p$ GS tokens shown in the lower branch
of Fig.~\ref{fig:framework2}. The proposal head is used to construct a
high-recall candidate pool rather than to determine the final token subset.

\subsubsection{Region-Aware Coreset Selection}
\label{sec:coreset_selection}

The provisional fine-token pool may contain candidates with nearby centers,
overlapping Gaussian support, or similar semantic content. The region-aware
coreset selector removes this redundancy while preserving the continuous field
represented by the full candidate pool. It retains all $K_c$ coarse tokens and
selects $K_f$ fine tokens from the $K_p$ provisional candidates, resulting in a
fixed final budget of $K=K_c+K_f$ tokens.

The selector uses two complementary forms of evidence, as illustrated in
Fig.~\ref{fig:framework2}. Region evidence estimates the semantic or anatomical
role of each candidate and prevents spatially small or underrepresented regions
from losing all token support. Field evidence measures how much each candidate
contributes to the continuous Gaussian field induced by the provisional token
set. During selection, candidates that improve uncovered regions receive higher
priority, whereas candidates that are spatially or geometrically redundant
with already selected tokens receive lower priority. The selected subset is
therefore optimized for joint coverage rather than independent token scores.

The proposal head and the coreset selector serve distinct purposes. The former
reduces the dense spatial search space and aims for high candidate recall before
the complete token geometry and semantics are available. The latter operates
on the constructed provisional GS tokens and performs set-level redundancy
reduction and fixed-budget compression using their learned geometry and region
evidence.

\subsubsection{Optional Local Prediction Refinement}
\label{sec:optional_refinement}

For ambiguous or uncertain regions, the base dense prediction can optionally
be enhanced by a lightweight local residual refiner. The refinement branch
identifies a small set of difficult locations using predictive uncertainty and,
when enabled, Gaussian-derived evidence such as token responsibility, coverage,
or semantic disagreement. It predicts a bounded residual correction only at
the selected locations and leaves the remaining prediction unchanged. This
branch does not alter the definition, construction, or fixed budget of the GS
tokens; it is therefore treated as an optional prediction enhancement rather
than a necessary component of the base GSToken architecture.

\section{Experiments}
\subsection{Experimental Setup}
We conduct experiments on the BraTS 2021 training set, from which we construct fixed development training and development validation splits while reserving 20 cases as an unused audit set. All results are reported on the same 246-case development validation split; neither the audit set nor the official test set is accessed.

We compare GSToken256 with the capacity-matched TokenLearner256 baseline and the higher-capacity Patch512 reference in terms of segmentation accuracy, boundary quality, lesion detection, and Transformer readability. All token representations are evaluated using the same feature-only Transformer reader architecture, comprising two Transformer encoder layers, six attention heads, and a feature dimension of 96. In the primary capacity-matched comparison, GSToken256 and TokenLearner256 each output 256 tokens with a serialized payload of 119,040 bytes, using identical data partitions, reader initialization, and training budgets. Patch512 instead outputs 512 tokens with a serialized payload of 238,080 bytes and is included solely as a higher-capacity reference. For the reported GSToken experiments, the importance scalar is fixed to \(\alpha_i=1\) for all tokens, and the learned importance head is disabled. The tokenizer and reader are each trained for 1,000 steps using AdamW with a learning rate of \(3\times10^{-4}\), and all experiments are repeated with random seeds 17, 23, and 41.

\subsection{Main Results}

\begin{figure}[htbp]
    \centering
    \includegraphics[width=\linewidth]{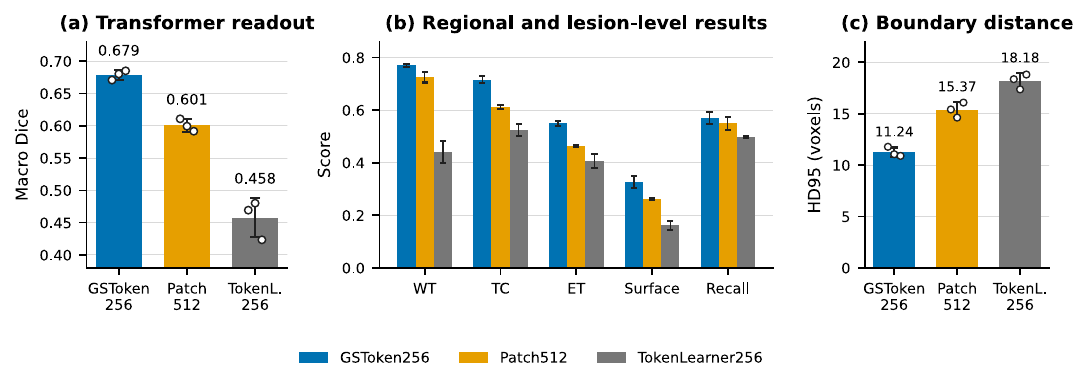}
    \caption{Cross-model comparison of the main experimental results.}
    \label{fig:main_results}
\end{figure}

We compare GSToken with Patch512 and TokenLearner256 under the frozen-token evaluation protocol. Figure~\ref{fig:main_results} summarizes the main quantitative results. Beyond these aggregate metrics, we also examine how each tokenizer distributes its spatial support over individual lesions.

As illustrated in Figure 4, the three tokenization strategies exhibit different spatial response patterns. Patch tokens produce rigid rectangular coverage, whereas TokenLearner yields concentrated but spatially fragmented responses. In contrast, GSToken produces smoother lesion-centered responses while retaining broader anatomical coverage. These visualizations illustrate how token semantics are distributed spatially, rather than serving as a direct quantitative measure of boundary accuracy.

\begin{figure}[htbp]
    \centering
    \includegraphics[width=\linewidth]{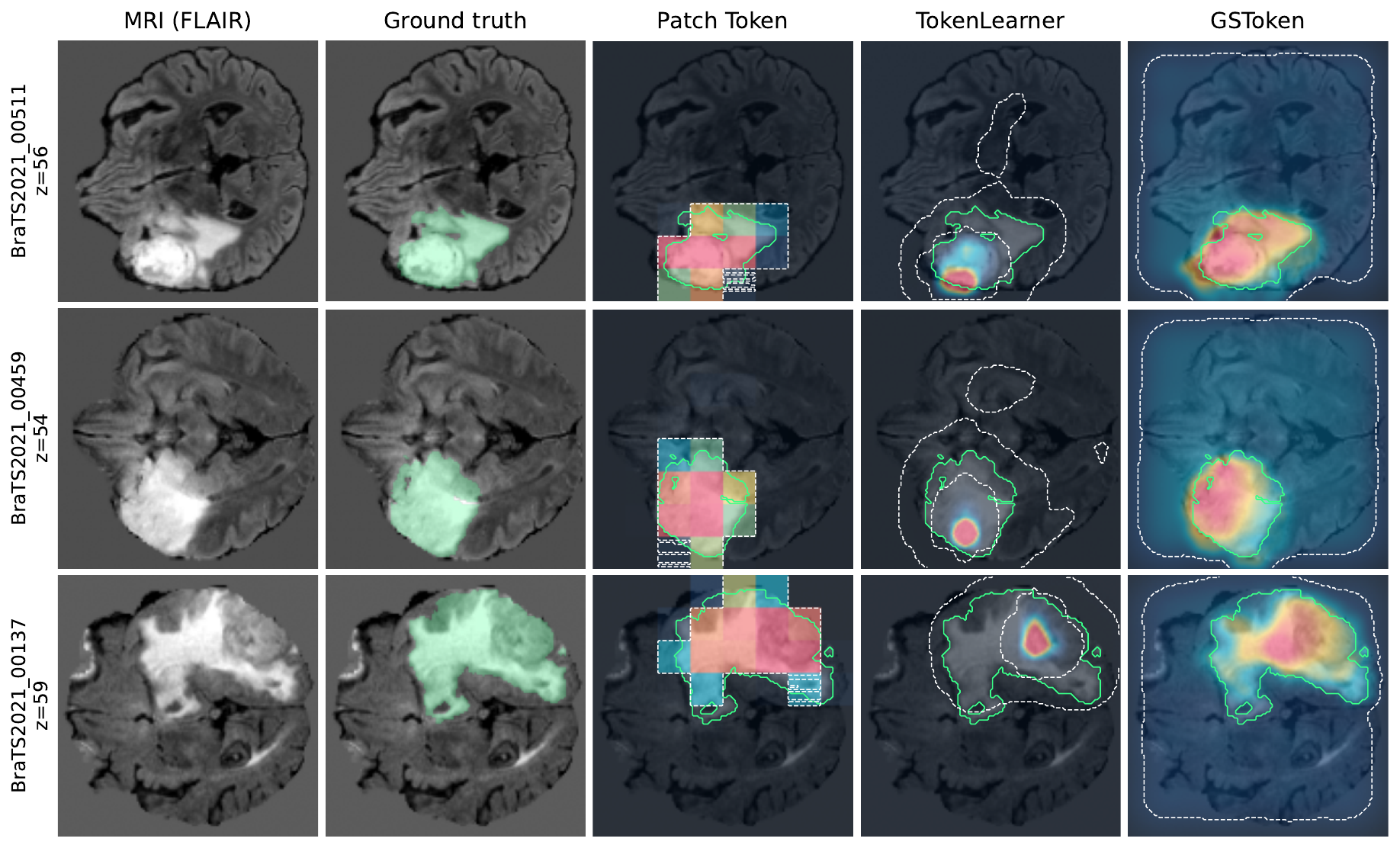}
    \caption{Token-to-image mapping visualization.}
    \label{fig:token_image_mapping}
\end{figure}

Quantitatively, GSToken employs a lesion-aware dynamic allocation strategy, combining 192 fine-grained tokens and 64 coarse-grained tokens to form a total of 256 tokens; Patch Token uses a regular grid partitioning scheme, while TokenLearner performs spatial aggregation via learnable attention. We repeated the experiment using three random seeds: 17, 23, and 41. As shown in Figure~\ref{fig:main_results}, GSToken’s average Macro Dice is \(0.6786\pm0.0073\), which is higher than that of the higher-capacity Patch512 (\(0.6005\pm0.0096\)) and the strictly capacity-matched TokenLearner256 (\(0.4576\pm0.0301\)). Compared to TokenLearner, GSToken achieves an average increase of 0.2210 in Macro Dice, 0.1645 in Surface Dice, and a reduction of 6.93 in HD95. This advantage holds across all three seeds, indicating that its explicit spatial support and dynamic allocation can preserve more three-dimensional tumor information readable by the Transformer within a fixed capacity. TokenLearner performed the weakest, suggesting that relying solely on attention aggregation may result in the loss of fine-grained spatial structures. In summary, GSToken demonstrates a stronger ability to preserve three-dimensional lesion information.

\section{Conclusion and Discussion}

We propose GSToken, a token representation with explicit spatial localization capabilities designed for compact 3D medical image modeling. GSToken employs learnable anisotropic Gaussian geometry to replace rigid patch-space anchoring or unconstrained attention aggregation, enabling each token to adaptively adjust its center position, spatial extent, and orientation based on the underlying anatomical structure. Furthermore, a coarse-to-fine allocation strategy combines global context coverage with local adaptive representation capabilities.
When using the same feature-only Transformer reader, GSToken256 consistently outperforms the capacity-matched TokenLearner256 and the higher-capacity Patch512 across three random seeds. The improvements achieved by GSToken on metrics such as Macro Dice, region segmentation, Surface Dice, lesion recall rate, and HD95 indicate that it not only preserves semantic information but also better retains the geometric structure of lesion extent and boundaries. Combined with token-to-image visualization results, these experiments demonstrate that, compared to fixed patch grids or purely attention-driven token aggregation, explicit continuous-space support can generate representations with stronger spatial consistency and greater sensitivity to lesion morphology.
Overall, GSToken is capable of preserving richer 3D lesion information in a compact form within a fixed token budget. The above results suggest that explicit Gaussian geometry holds promise as an effective alternative to traditional tokenization methods in volumetric medical image analysis.

However, compared to established strong segmentation baselines such as nnU-Net, the current GSToken model still lags behind in segmentation accuracy. The experiments in this paper primarily validate the ability of Gaussian Tokens to preserve three-dimensional lesion information within a fixed token budget, rather than pursuing optimal performance for a complete segmentation system. Future research will focus on more efficient learning of Gaussian geometric parameters, cross-scale feature fusion, and token allocation mechanisms, as well as exploring the joint optimization of GSToken with more powerful encoders and decoders to fully leverage its spatial representation advantages and improve practical segmentation performance.

\medskip
{\small
\bibliographystyle{unsrt}   % 按引用顺序编号
\bibliography{ref}          % 读入 ref.bib
}

\clearpage
\appendix

\section{Dataset Split and Preprocessing}
\label{app:data}

\paragraph{Dataset split.}
We use the 1,251 labeled cases from the BraTS 2021 training set. Twenty cases are first reserved as an audit split and are not accessed during model development or evaluation. The remaining 1,231 cases are deterministically shuffled using seed 17 and divided into 985 development-training cases and 246 development-validation cases. The same case manifests are reused for every tokenizer and for all experimental seeds. The development-validation split is used for checkpoint selection and comparative reporting. Neither the reserved audit split nor the official BraTS test set is accessed in the experiments reported in this paper.

\paragraph{Image preprocessing.}
Each case contains four MRI modalities: T1, contrast-enhanced T1 (T1ce), T2, and FLAIR. Each modality is independently z-score normalized over its nonzero voxels, while background voxels remain zero. We compute a common brain bounding box from the union of the nonzero regions across all modalities and extend it by a margin of three voxels in each spatial direction. The four modalities and the segmentation mask are then cropped using the same bounding box.

The original BraTS enhancing-tumor label 4 is remapped to class index 3. The resulting class indices are background (0), necrotic and non-enhancing tumor core (1), peritumoral edema (2), and enhancing tumor (3). Following the BraTS convention, whole tumor (WT), tumor core (TC), and enhancing tumor (ET) are defined as
\[
\mathrm{WT}=\{1,2,3\}, \qquad
\mathrm{TC}=\{1,3\}, \qquad
\mathrm{ET}=\{3\}.
\]
After converting the NIfTI spatial order to the depth-height-width convention used by 3D convolutions, all images are resized to \(128^3\) using trilinear interpolation. Segmentation masks are resized using nearest-neighbor interpolation. During tokenizer training, independent random flips are applied along each of the three spatial axes with probability 0.5. No data augmentation is applied during validation or frozen-token evaluation.

\section{Serialized Token Contract}
\label{app:contract}

To compare different tokenization strategies under a common capacity definition, every tokenizer exports a model-independent token packet. For a token budget \(K\), the packet contains the fields listed in Table~\ref{tab:packet-contract}. All continuous fields are stored in 32-bit floating-point format, and the active-token mask is stored as a Boolean value requiring one byte per token.

\begin{table}[t]
\centering
\small
\caption{Fields in the serialized token packet. ``Reader input'' indicates whether a field is provided to the frozen feature-only Transformer reader.}
\label{tab:packet-contract}
\begin{tabular}{lccc}
\hline
Field & Values per token & Data type & Reader input \\
\hline
Content feature & 96 & float32 & Yes \\
Center & 3 & float32 & Yes \\
Anisotropic scale & 3 & float32 & Yes \\
Rotation matrix & 9 & float32 & Yes \\
Importance & 1 & float32 & Yes \\
Semantic logits & 4 & float32 & No \\
Active mask & 1 & bool & Yes \\
\hline
\end{tabular}
\end{table}

The complete serialized payload is therefore
\[
\begin{aligned}
B(K)
&=
K\left[(96+3+3+9+1+4)\times4+1\right] \\
&=465K \quad \text{bytes}.
\end{aligned}
\]
Consequently, GSToken256 and TokenLearner256 each require
\[
B(256)=119{,}040 \quad \text{bytes},
\]
whereas the higher-capacity Patch512 reference requires
\[
B(512)=238{,}080 \quad \text{bytes}.
\]

Semantic logits are retained in the serialized packet to preserve a uniform packet format and to support post-hoc visualization, but they are explicitly excluded from the feature-only reader. The reader-visible portion of a 256-token packet is therefore
\[
256\left[(96+3+3+9+1)\times4+1\right]
=114{,}944 \quad \text{bytes}.
\]
Because the same exclusion is applied to every tokenizer, it does not introduce a method-specific side channel or capacity advantage.

For the reported GSToken experiments, the importance scalar is fixed to \(\alpha_i=1\) for every token, and the learned importance head is disabled. The importance field remains in the packet to maintain the same contract across tokenization methods.

\section{Frozen-Token Evaluation and Statistical Analysis}
\label{app:frozen-probe}

After each tokenizer is trained for 1,000 steps, all tokenizer parameters are frozen and token packets are exported for the development-training and development-validation cases without gradient computation. A separate feature-only Transformer reader is then trained from scratch for each tokenizer using the same architecture, seed-matched initialization, data partition, and optimization budget. The reader receives only the token features, centers, scales, rotations, importance values, and active masks; it has no access to the original MRI volume, dense encoder features, encoder skip connections, or token-level semantic logits. Each reader is trained for 1,000 steps using AdamW with a learning rate of \(3\times10^{-4}\) and a weight decay of \(10^{-5}\).

All metrics are computed independently for the same 246 development-validation cases. Macro Dice is the arithmetic mean of the WT, TC, and ET Dice scores. Within each seed, paired differences between GSToken and the comparison method are calculated using identical case identifiers. Paired 95\% confidence intervals are estimated using 20,000 bootstrap resamples of the case-level differences with replacement. Experiments are repeated using seeds 17, 23, and 41, and the final results are reported as the mean and sample standard deviation across seeds. Any Student-\(t\) interval based on the three seeds is treated as descriptive only.

%\newpage
%\input{checklist.tex}

\end{document}